\documentclass[final]{cvpr}

\usepackage{times}
\usepackage{epsfig}
\usepackage{graphicx}
\usepackage{amsmath}
\usepackage{amssymb}

\usepackage{enumitem}
\usepackage[ruled]{algorithm2e}
\usepackage{array}
\usepackage{multirow}
\usepackage{color}
\usepackage{cases}
\usepackage[export]{adjustbox}
\usepackage[dvipsnames]{xcolor}
\usepackage{textpos}
\usepackage{comment}
\usepackage{textcomp}
\usepackage{listings}
\usepackage{bibentry}

\usepackage{rotating}

\usepackage{booktabs}
\newcommand{\B}{\bf}

\usepackage[pagebackref=true,breaklinks=true,colorlinks,bookmarks=false]{hyperref}

\def\cvprPaperID{1} 
\def\confYear{CVSports 2021}

\newcolumntype{C}[1]{>{\centering\let\newline\\\arraybackslash\hspace{0pt}}b{#1}}

\begin{document}

\title{Temporally-Aware Feature Pooling for Action Spotting in Soccer Broadcasts}

\author{Silvio Giancola \quad \quad Bernard Ghanem \\
King Abdullah University of Science and Technology, Saudi Arabia\\
{\tt\small \{silvio.giancola,bernard.ghanem\}@kaust.edu.sa}
}

\maketitle
\thispagestyle{empty}

\newcommand{\mysection}[1]{\vspace{2pt}\noindent\textbf{#1}}
\newcommand{\Table}[1]{Table~\ref{tab:#1}}
\newcommand{\Figure}[1]{Figure~\ref{fig:#1}}
\newcommand{\Equation}[1]{Equation~\eqref{eq:#1}}
\newcommand{\Equations}[2]{Equations \eqref{eq:#1} and \eqref{eq:#2}}
\newcommand{\Section}[1]{Section~\ref{sec:#1}}
\newcommand{\SoccerNet}{SoccerNet~\cite{Giancola_2018_CVPR_Workshops}\xspace}
\newcommand{\SN}[1]{SoccerNet-v{#1}\xspace}
\newcommand{\sota}{state-of-the-art\xspace}
\newcommand{\Sota}{State-of-the-art\xspace}
\newcommand{\SotA}{State-of-the-Art\xspace}
\newcommand{\SOTA}{State-Of-The-Art\xspace}
\newcommand{\CALF}{CALF~\cite{cioppa2020context}\xspace}
\newcommand{\RMS}{RMS-Net~\cite{tomei2021rms}\xspace}
\newcommand{\ActivityNet}{ActivityNet~\cite{caba2015activitynet}\xspace}
\newcommand{\NetVLAD}{NetVLAD~\cite{arandjelovic2016netvlad}\xspace}
\newcommand{\VLAD}{VLAD~\cite{jegou2010aggregating}\xspace}
\newcommand{\cl}{\mathbf{c}}
\newcommand{\x}{\mathbf{x}}
\newcommand{\w}{\mathbf{w}}
\newcommand{\X}{\mathbf{X}}
\newcommand{\Vlad}{\mathbf{V}}

\newcommand{\TODO}[1]{\textcolor{red}{[TODO:#1]}}
\newcommand{\SG}[1]{\textcolor{red}{[SG:#1]}}

\newcommand\blfootnote[1]{%
  \begingroup
  \renewcommand\thefootnote{}\footnote{#1}%
  \addtocounter{footnote}{-1}%
  \endgroup
}


\begin{abstract}
    Toward the goal of automatic production for sports broadcasts, a paramount task consists in understanding the high-level semantic information of the game in play. For instance, recognizing and localizing the main actions of the game would allow producers to adapt and automatize the broadcast production, focusing on the important details of the game and maximizing the spectator engagement. In this paper, we focus our analysis on action spotting in soccer broadcast, which consists in temporally localizing the main actions in a soccer game. To that end, we propose a novel feature pooling method based on NetVLAD, dubbed \textbf{NetVLAD++}, that embeds temporally-aware knowledge. Different from previous pooling methods that consider the temporal context as a single set to pool from, we split the context \emph{before} and \emph{after} an action occurs. We argue that considering the contextual information around the action spot as a single entity leads to a sub-optimal learning for the pooling module. With NetVLAD++, we disentangle the context from the \emph{past} and \emph{future} frames and learn specific vocabularies of semantics for each subsets, avoiding to blend and blur such vocabulary in time. Injecting such prior knowledge creates more informative pooling modules and more discriminative pooled features, leading into a better understanding of the actions. We train and evaluate our methodology on the recent large-scale dataset \SN2, reaching $53.4\%$ Average-mAP for action spotting, a $+12.7\%$ improvement w.r.t the current \sota.
\end{abstract}


\section{Introduction}
\label{sec:Intro}

\begin{figure}
    \centering
    \includegraphics[width=\linewidth]{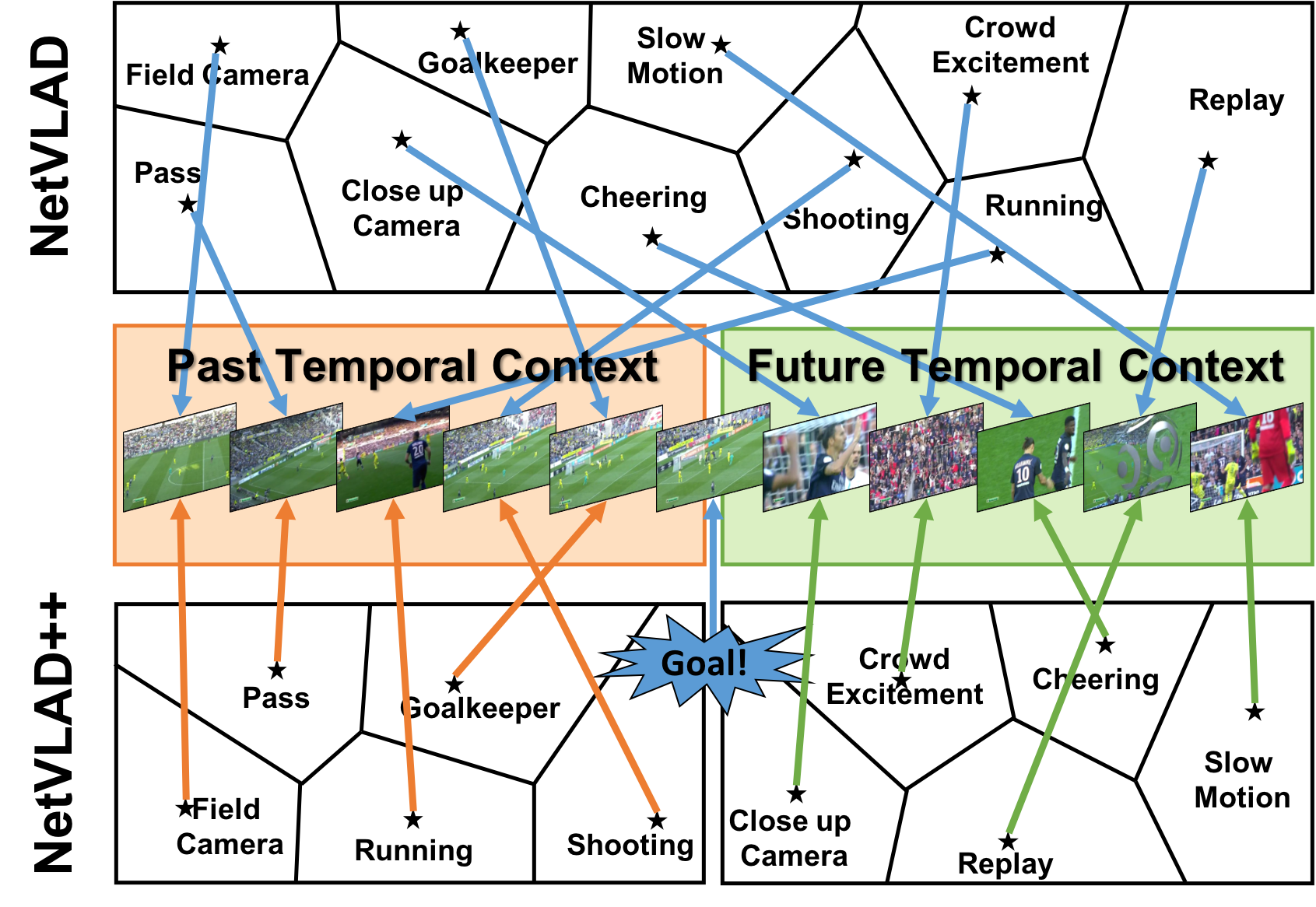}
    \caption{
    \textbf{NetVLAD} (top) \vs \textbf{NetVLAD++} (bottom) pooling modules for action spotting. 
    Our temporally-aware NetVLAD++ pooling module learns specific vocabularies for the \textit{past} and \textit{future} semantics around the action to spot.
    }
    \label{fig:Pooling}
\end{figure}

The volume of sports TV broadcast available worldwide increased at a fast pace over the last years. The amount of hours of sports TV broadcasted in the United States from 2002 to 2017 has grown ${>}4\times$~\cite{SportsBroadcastUS}, with similar trends in European countries~\cite{VolumeSportsFrance,VolumeSportsFrancePay}. Consequently, the market size for the sports media rights is booming~\cite{SportsMediaRightsUS}, revealing a ludicrous market estimated to worth ${>}25$B USD by 2023 in US alone. Yet, creating broadcast contents still requires a tremendous manual effort from the producers who could heavily benefit from automated processes.

Autonomous broadcast production requires an understanding of the sports it focuses on. In particular, it needs to be aware of \textit{where} to look at, \eg focusing the camera on spatial areas of the field, but also \textit{when} to look at, \eg focusing on a given actions of interest temporally anchored in the broadcast. While most literature focuses on \textit{where} to look at, less effort were put on \textit{when} to look at. As an example, player or ball detection and tracking algorithms reach excellent performances \cite{manafifard2017survey,Sarkar_2019_CVPR_Workshops,Cioppa_2019_CVPR_Workshops, bridgeman2019multi} and are commonly used as priors to identify \textit{where} to look at, by simply regressing the extrinsic parameters of the cameras to center those objects \cite{thomas2017computer}. Yet, the level of semantics involved in this task is rather low and does not require higher understanding of the game. On the other hand, identifying \textit{when} to look at requires understanding higher semantics level, closer to the game, focusing on abstract action concepts rather than well defined actor's or object's priors. To that end, we believe that understanding actions rather than actors is a more challenging task, yet to explore in sports videos.

In this work, we propose to tackle the task of action spotting, \ie localizing well-defined actions in time, anchored with single timestamps along a video. Such task is commonly solved by looking at a temporal context around a given timestamp and regressing the actionness in time. The class-aware actionness is then reduced to a singularly anchored spot using non maximum suppression techniques \cite{Giancola_2018_CVPR_Workshops}. Previous works on that realm consider temporal pooling techniques~\cite{Giancola_2018_CVPR_Workshops,Vanderplaetse2020Improved}, spatio-temporal encoding~\cite{rongved-ism2020}, multi-tower temporal CNN~\cite{vats2020event} or context-aware regression modules~\cite{cioppa2020context}. Inspired by those related works, we propose a temporally-aware pooling technique that push forward the previous \sota. In particular, we developed a pooling module that consider the near \textit{past} and \textit{future} context around the action, independently. Our novel temporal module, dubbed \textbf{NetVLAD++}, is based on two NetVLAD pooling layer across the frames \textit{before} and \textit{after} the action occurs, respectively. Such temporal awareness brings a significant boost in the performances in the SoccerNet-v2 benchmark, leading into \sota performances for action spotting.

\mysection{Contributions.} We summarize them as follow:
    \textbf{(i)} 
    We introduce NetVLAD++, a novel pooling module for action spotting that learns a temporally-aware vocabulary for past and future temporal context. 
    \textbf{(ii)} 
    We implement a more efficient architecture for action spotting, in term of memory and computational complexity, leading to \sota performances for action spotting on \SN2.
    \textbf{(iii)} 
    We propose a comprehensive ablation that points out the contributions of each architectural block design.

%
%
%

\section{Related Work}
\label{sec:SOTA}



\mysection{Computer Vision in Soccer.}
The literature in soccer-related computer vision mainly focuses on low-level understanding of a soccer broadcast~\cite{moeslund2014computer}, \eg
localizing a field and its lines~\cite{Cioppa2018ABottom,farin2003robust,homayounfar2017sports}, detecting players~\cite{Cioppa_2019_CVPR_Workshops,yang2017robust}, their motion~\cite{felsen2017will,manafifard2017survey}, their pose~\cite{Bridgeman_2019_CVPR_Workshops, Zecha_2019_CVPR_Workshops}, their team~\cite{Istasse_2019_CVPR_Workshops}, the ball~\cite{Sarkar_2019_CVPR_Workshops,Theagarajan_2018_CVPR_Workshops} or a pass feasibility~\cite{Sangesa2020UsingPB}. Understanding frame-wise information is useful to enhance the visual experience of sports viewers~\cite{Rematas_2018_CVPR} and to gather player statistics~\cite{thomas2017computer}, but it falls short of higher-level game understanding needed for automatic editing purposes.
With the appearance of large scale datasets such as \SoccerNet, Yu~\etal~\cite{yu2018comprehensive} and SoccerDB~\cite{Jiang2020SoccerDB}, higher level tasks started to appear. \SoccerNet introduced the task of action spotting, \ie localizing every action with its timestamp in a large corpus of TV broadcasts. They introduced a dataset of $500$ games from the European leagues, annotated with $6637$ actions of goals, cards and substitutions. Yu~\etal~\cite{yu2018comprehensive} released a novel dataset of $222$ broadcast videos of $45$ min each. They introduced interesting annotations of camera shots, players position, events and stories, yet do not provide any task nor baseline on how to use those annotations. SoccerDB~\cite{Jiang2020SoccerDB} merged a subset of $270$ games from SoccerNet with $76$ soccer games from the Chinese Super League. They proposed several tasks, ranging from object detection, action recognition, temporal action localization and replay segmentation. Lastly, \SN2~\cite{SNv2} extended \SoccerNet with more than $300k$ extra annotations and propose novel tasks that would support the automatic production of soccer broadcast. In particular, \SN2~\cite{SNv2} extended the task of action spotting to $17$ classes to understand the fine-grained details of a soccer game. They also introduced two novel tasks: camera shot segmentation for broadcast editing purposes and replay grounding for highlight and summarization purposes. In this work, we leverage the fine-grained annotations from \SN2~\cite{SNv2} and compete in the task of action spotting.

\mysection{Action spotting.}
Action spotting was introduced in \SoccerNet and defined as the localization of a instantaneous event anchored with a single timestamp, namely an \textit{action}, in contrast with \textit{activities}, defined with a start and an end~\cite{caba2015activitynet}. It draws similarities with the concept of action completion~\cite{heidarivincheh2017detecting} where an action is defined with a single anchor in time, but serve a different purpose of predicting the future completion of that action. Giancola~\etal~\cite{Giancola_2018_CVPR_Workshops} introduced a first baseline on \SoccerNet based on different pooling techniques. Yet, their code is hardly optimized, leading into an strong under-estimation of the pooling performances for action spotting. Vanderplaetse~\etal~\cite{Vanderplaetse2020Improved} later improved that baseline by merging visual and audio features in a multi-modal pooling approach. Rongved~\etal~\cite{rongved-ism2020} trained a 3D ResNet encoder~\cite{tran2018closer} directly from the video frames. Although the performances were far from the baseline, mostly due to the difficulty of training an encoder from scratch, the technical prowess lied in training end-to-end for action spotting with $16$ V100 GPU combining $512$GB of memory. Vats~\etal~\cite{vats2020event} leveraged a multi-tower CNN to process information at various temporal scales to account for the uncertainty of the action locations. 
Cioppa~\etal~\cite{cioppa2020context} proposed a method based on a context-aware loss function that model the temporal context surrounding the actions. They propose an alternative approach that predicts multiple spots from each chunk of video, by regressing multiple temporal offsets for the actions.
Most recently, Tomei~\etal~\cite{tomei2021rms} introduced a regression and masking approach with RMS-Net, inspired by common detection pipeline \cite{ren2016faster} and self-supervised pre-training~\cite{devlin2018bert}.
It is worth noting that they reach impressive performance gains by fine tuning the last ResNET block of the video frame encoder.
In our work, we improve the original temporal pooling mechanism proposed in \SoccerNet by introducing temporally-aware bag-of-words pooling modules. 
Unlike Tomei~\etal~\cite{tomei2021rms}, we refrain from fine-tuning the pre-extracted ResNET frame features for a fair comparison with the related work, but simply allow for a learnable projection (similar to PCA) to reduce the feature dimensionality.


\section{Methodology}
\label{sec:Dataset}

In this section, we first recall the definition of NetVLAD and propose a more computationally efficient implementation (\ref{subsec:netvlad}), we present our novel temporally-aware NetVLAD pooling module learning the \textit{past} and \textit{future} temporal context independently (\ref{subsec:netvlad++}) and its implementation in a more comprehensive pipeline for action spotting (\ref{subsec:network}).

\subsection{Recall on NetVLAD}
\label{subsec:netvlad}

\NetVLAD is a differentiable pooling technique inspired by \VLAD. In particular, VLAD learns clusters of features descriptors and defines an aggregation of feature as the average displacement of each features with respect to the center of its closer cluster. NetVLAD generalizes VLAD by \textbf{(i)} softening the assignment for full-differentiable capability, and \textbf{(ii)} disentangling the definition of the cluster and the assignment of the samples.

\mysection{VLAD.} Formally, given a set of $N$ $D$-dimensional features $\{\x_i\}_{i=1..N}$ as input, a set of $K$ clusters centers $\{\cl_k\}_{k=1..K}$ with same dimension $D$ as VLAD parameters, the output of the VLAD descriptor $V$ is defined by:

\begin{equation}
    V(j,k) = \sum_{i=1}^N a_k(\x_i)(\x_i(j)-\cl_k(j))
    \label{eq:vlad}
\end{equation}

where $\x_i(j)$ and $\cl_k(j)$ are respectively the $j$-th dimensions of the $i$-th descriptor and $k$-th cluster center. $a_k(\x_i)$ denotes the hard assignment of the sample $\x_i$ from its closer center, \ie $a_k(\x_i)=1$ if $\cl_k$ is the closest center of $\x_i$, $0$ otherwise. The matrix $V$ is then L2-normalized at the cluster level, flatten into a vector of length $D \times K$ and further L2-normalized globally.

\mysection{NetVLAD.} The VLAD module is non-differentiable due to the hard assignment $a_k(\x_i)$ of the samples $\{\x_i\}_{i=1}^N$ to the clusters $\{\cl_k\}_{i=1}^K$. Those hard-assignment creates discontinuities in the feature space between the clusters, impeding gradients to flow properly. To circumvent this issue, \NetVLAD introduces a soft-assignment $\Tilde{a}_k(\x_i)$ for the samples $\{\x_i\}_{i=1}^N$, based on their distance to each cluster center. Formally:

\begin{equation}
    \Tilde{a}_k(\x_i) = \frac { e^{-\alpha \| \x_i - \cl_k\|^2} } { \sum_{k'=1}^K e^{-\alpha \| \x_i - \cl_{k'}\|^2} }
    \label{eq:softassignment}
\end{equation}

$\Tilde{a}_k(\x_i)$ ranges between $0$ and $1$, with the highest value assigned to the closest center. $\alpha$ is a temperature parameter that controls the softness of the assignment, a high value for $\alpha$ (\eg $\alpha \xrightarrow[]{} + \infty$) would lead to a hard assignment like in VLAD. Furthermore, by expanding the squares and noticing that $ e^{-\alpha \| \x_i \|^2} $ will cancel between the numerator and the denominator, we can interpret Equation~\eqref{eq:softassignment} as the softmax of a convolutional layer for the input features parameterized by $\w_k = 2 \alpha \cl_k$ and $b_k = - \alpha \| \cl_k \| ^2$. Formally:

\begin{equation}
    \Tilde{a}_k(\x_i) = \frac { e^{\w_k^T \x_i + b_k} } { \sum _{k'} e^{\w_{k'}^T \x_i + b_{k'}} }
    \label{eq:convsoftmax}
\end{equation}

Finally, by plugging the soft-assignment from \eqref{eq:convsoftmax} into the VLAD formulation in \eqref{eq:vlad}, the NetVLAD features are defined as in Equation \eqref{eq:netvlad}, later L2-normalized per cluster, flattened and further L2-normalized in its entirety.

\begin{equation}
    V(j,k) = \sum_{i=1}^N\frac { e^{\w_k^T \x_i + b_k} } { \sum _{k'} e^{\w_{k'}^T \x_i + b_{k'}} } (\x_i(j)-\cl_k(j))
    \label{eq:netvlad}
\end{equation}

Note that the original VLAD optimizes solely the cluster centers $\cl_k$, while NetVLAD optimizes for $\{\w_k\}$, $\{b_k\}$ and $\{\cl_k\}$ independently, dropping the constraint of $\w_k = 2 \alpha \cl_l$ and $b_k = - \alpha \| \cl_k \| ^2$. These constraints were similarly dropped in \cite{arandjelovic2013all}, arguing for further freedom in the training process.

\mysection{Efficient implementation.} Implementing NetVLAD with libraries such as Tensorflow or Pytorch could lead to several memory challenges in mini-batch training. In particular, the formulation in \eqref{eq:netvlad} would lead to a 4-dimensional tensor, in particular due to the residuals $(\x_i(j)-\cl_k(j))$, defined with a  batch size ($B$), a set size ($N$), a number of clusters ($K$) and a features dimension ($D$). With small considerations in Equation~\eqref{eq:netvlad}, in particular splitting the residual in the two operands like in Equation~\eqref{eq:netvladloupe}, leads to a difference of two 3D tensors only, reducing the memory footprint as well as the computational complexity. Empirically, we experienced a ${\sim}5 \times$ speed up in computation (backward and forward) and a similar reduction for the memory footprint.

\begin{equation}
    \begin{split}
        V(j,k) &= \sum_{i=1}^N \Tilde{a}_k(\x_i) ( \x_i(j) - \cl_k(j) ) \\
               &= \sum_{i=1}^N \Tilde{a}_k(\x_i) \x_i(j) - \Big(\sum_{i=1}^N \Tilde{a}_k(\x_i)\Big)\cl_k(j )
    \end{split}
    \label{eq:netvladloupe}
\end{equation}

\mysection{Pooling for Action Spotting.} 
We follow a similar architecture structure proposed in \SoccerNet.
In particular, we learn to classify whether specific actions occurs within a temporal window. For inference, we densely slide the temporal window along the video to produce a class-aware actionness, on top of which we apply a non-maximum suppression (NMS). The frame features are pre-computed and pre-reduced in dimension with PCA, then pooled along the sliding window to predict the actionness of the central frame. Yet, \SoccerNet does not optimize the dimensionality reduction for the end-task and the pooling method is not aware of the temporal order of the frame features, nor consider \textit{past} and \textit{future} context independently.

\subsection{NetVLAD++: Temporally-aware pooling}
\label{subsec:netvlad++}

We propose a temporally-aware pooling module dubbed NetVLAD++ as our primary contribution. The VLAD and NetVLAD pooling methods are permutation invariant, as a consequence, do not consider the order of the frames, but only aggregates the features as a set. In the particular case of action spotting, the frames features from the videos are temporally ordered in time, and can be categorized between \textit{past} and \textit{future} context.

As noted by Cioppa \etal \cite{cioppa2020context}, the amount of context embedded before and after an action occurs is different, yet complementary. In addition, we argue that different actions might share similar vocabulary either before or after those actions occur, but usually not both. As an example, the semantic information contained \textit{before} a ``goal'' occurs and \textit{before} a ``shot on/off target'' occurs are similar, representing a lower level semantic concept of a player shooting on a target and a goalkeeper trying to catch that ball. Yet, those two action classes depict different contextual semantics \textit{after} it occurs, with the presence of cheering (for ``goal'') or frustration (for simple ``shot'') in the players. Following a similar logic, the spotting of a ``penalty'' would benefit more from the knowledge of what happened \emph{before} that penalty was shot, as the follow-up cheering would look similar to any other goal. Without loss of generality, it appears that the amount of information to pool among the features \emph{before} and \emph{after} an action occurs might contain different low-level semantics, helping identifying specific fine-grained actions.

To that end, we propose a novel temporally-aware pooling module, dubbed \textbf{NetVLAD++}, as depicted in Figure~\ref{fig:Pooling}. In particular, we learn $2$ different NetVLAD pooling modules for the frame features from \textit{before} and \textit{after} an action occurs. We define the \textit{past} context as the frame feature with a temporal offset in $[-T_b, 0[$ and the \textit{future} context as the frame feature with a temporal offset in $[0, T_a]$. Each pooling module aggregates different clusters of information from the $2$ subsets of features, using $K_a$ and $K_b$ clusters, respectively for the \textit{after} and \textit{before} subsets. Formally:

\begin{equation}
\begin{split}
    V = \Box( V_b, V_a )
\end{split} 
    \label{eq:netvlad++}
\end{equation}

with $\Box$ an aggregation of $V_b$ and $V_a$ that represent the NetVLAD pooled features for the sample \textit{before} and \textit{after} the action occurs, parameterized with $K_b$ clusters for the \textit{past} context and $K_a$ clusters for the \textit{future} context.

\subsection{Architecture for Action Spotting}
\label{subsec:network}

\begin{figure}
    \centering
    \includegraphics[width=\linewidth]{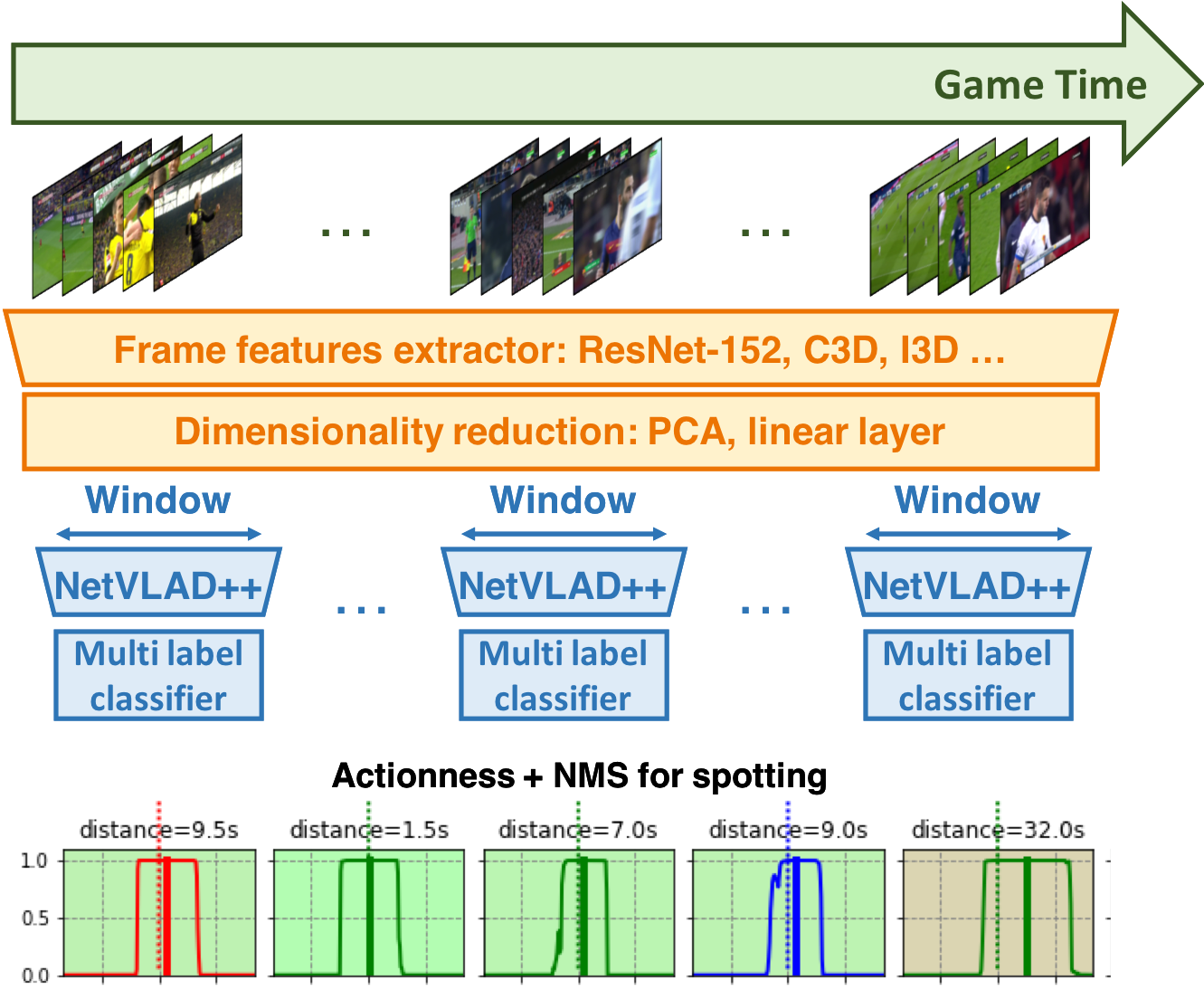}
    \caption{Action spotting architecture based on our novel temporally aware pooling module.}
    \label{fig:architecture}
\end{figure}

We integrated our novel pooling module into a larger architecture depicted in Figure \ref{fig:architecture}, follows a similar structure presented in \SoccerNet. In particular, it is based on a pre-trained frame feature encoder, a dimensionality reduction, a pooling module from a temporally sliding window and a per-frame classifier that depicts a class-aware actionness. The action spotting is then performed using a non-maximum suppression (NMS). The main difference with \SoccerNet are twofold: an end-to-end learnable dimensionality reduction layer different from PCA and a temporally-aware pooling module.

\mysection{Video encoding.} We use the features extracted by \SN2, based on ResNet-152 \cite{he2016deep} pre-trained on ImageNet \cite{deng2009imagenet}. The weights are frozen and the frame features are pre-extracted at 2fps with a resolution of 224x224, scaled down in height then cropped on the sides for the width. The features correspond to the activation of the last layer of the ResNet-152 architecture, after the max pooling across the 2D feature map and before the classification layer, resulting in features of dimension $2048$. We consider those features as input for the remaining of the architecture.

\mysection{Dimensionality reduction.} The dimension of the features are reduced from $2048$ to $512$, following \SoccerNet that learned a PCA reduction to that dimension. We argue that a linear layer would learn a better linear combination of the frame features, by removing the orthogonality constraint introduced by PCA. We refer to the experiments to appreciate the boost in performances. Moreover, learning a PCA reduction is feasible offline only, hence not practical for online training as it require the feature to be pre-extracted.

\mysection{Temporally-aware pooling.} We consider window chunks of time $T$ s along the video. The temporally contiguous set of features are split equally \textit{before} and \textit{after} the center of the window and pooled accordingly. We normalize the features along the feature dimension and apply the $2$ NetVLAD module for each subset of features. The $2$ output NetVLAD features are concatenated along the feature dimension, leading into a feature of dimension $(K_b + K_a) \times D$.

\begin{table*}[ht]
\scriptsize
    \caption{\textbf{\Sota comparison.}
    We report the results of NetVLAD++ for action spotting (Average-mAP \%) on \SN2~\cite{SNv2}. 
    We report the performances of our best model over $5$ runs and detail its performances for each action class.
    }
    \centering
    \setlength{\tabcolsep}{2pt}
    \resizebox{\linewidth}{!}{
    \begin{tabular}{l||c||c|c||c|c|c|c|c|c|c|c|c|c|c|c|c|c|c|c|c}
     &  \begin{turn}{90}\bf SoccerNet-v2\end{turn} &  \begin{turn}{90}visible\end{turn}   &  \begin{turn}{90}unshown\end{turn}  & \begin{turn}{90} Ball out \end{turn} & \begin{turn}{90}Throw-in\end{turn} & \begin{turn}{90}Foul \end{turn} & \begin{turn}{90}Ind. free-kick \end{turn} & \begin{turn}{90}Clearance \end{turn} & \begin{turn}{90}Shots on tar. \end{turn} & \begin{turn}{90}Shots off tar. \end{turn} & \begin{turn}{90}Corner \end{turn} & \begin{turn}{90}Substitution \end{turn} & \begin{turn}{90}Kick-off \end{turn} & \begin{turn}{90}Yellow card \end{turn} & \begin{turn}{90}Offside \end{turn} & \begin{turn}{90}Dir. free-kick \end{turn} & \begin{turn}{90}Goal \end{turn} & \begin{turn}{90}Penalty \end{turn} & \begin{turn}{90}Yel.$\to$Red \end{turn} & \begin{turn}{90}Red card\end{turn} \\ 

       \midrule \midrule
\B MaxPool~\cite{Giancola_2018_CVPR_Workshops}   &  18.6 &  21.5 &  15.0 &  38.7 &  34.7 &  26.8 &  17.9 &  14.9 &  14.0 &  13.1 &  26.5 &  40.0 &  30.3 &  11.8 &   2.6 &  13.5 &  24.2 &   6.2 &  0.0 &  0.9 \\ \midrule
\B NetVLAD~\cite{Giancola_2018_CVPR_Workshops}   &  31.4 &  34.3 &  23.3 &  47.4 &  42.4 &  32.0 &  16.7 &  32.7 &  21.3 &  19.7 &  55.1 &  51.7 &  45.7 &  33.2 &  14.6 &  33.6 &  54.9 &  32.3 &  0.0 &  0.0 \\ \midrule

\B AudioVid~\cite{Vanderplaetse2020Improved}   &  39.9 &  43.0 &  23.3 &  54.3 &  50.0 &  55.5 &  22.7 &  46.7 &  26.5 &  21.4 &  66.0 &  54.0 &  52.9 &  35.2 &  24.3 &  46.7 &  69.7 &  52.1 &  0.0 &  0.0 \\ \midrule
\B CALF~\cite{cioppa2020context}               &  40.7 &  42.1 &  29.0 &  63.9 &  56.4 &  53.0 &  41.5 &  51.6 &  26.6 &  27.3 &  71.8 &  47.3 &  37.2 &  41.7 &  25.7 &  43.5 &\B72.2 &  30.6 &  0.7 &  0.7 \\ \midrule \midrule 

\B NetVLAD++                                   &\B53.4 &\B59.4 &\B34.8 &\B70.3 &\B69.0 &\B64.2 &\B44.4 &\B57.0 &\B39.3 &\B41.0 &\B79.7 &\B68.7 &\B62.1 &\B56.7 &\B39.3 &\B57.8 &  71.6 &\B79.3 &\B3.7 &\B4.0  \\ \bottomrule
    \end{tabular}}
    \label{tab:MainResults}
\end{table*}

\mysection{Video Chunk Classification.} In training, we consider non-overlapping window chunks with a sliding window of stride $T$. We build a classifier on top of the pooled feature, composed of a single neural layer with sigmoid activation and dropout. Since multiple actions can occur in the same temporal window, we consider a multi-label classification approach. A video chunk is labeled with all classes that appear on the chunk with a multi-label one-hot encoding. Similar to \SoccerNet, we optimize for a multi-label binary cross-entropy loss as defined in Equation \eqref{eq:loss}.

\begin{equation}
    \mathcal{L} = \frac{1}{N} \sum_{i=1}^N  y_i \log{ ( x_n ) } +  ( 1 - y_i ) \log{ ( 1 - x_n ) }
    \label{eq:loss}
\end{equation}

\mysection{Inference.} We run the sliding window of time $T$ along unseen videos with a temporal stride of $1$ to report the class-aware actionness scores in time. Similar to \cite{Giancola_2018_CVPR_Workshops,cioppa2020context,SNv2}, we use a Non Maximum Suppression (NMS) module to reduce positive spots closer than a given temporal threshold $T_{NMS}$. 



\section{Experiments}
\label{sec:Exp}


\mysection{Architectural design.} We set our temporally-aware NetVLAD pooling to have as many parameters and similar complexity as a traditional NetVLAD pooling. We refrain on adapting the size of the vocabulary for the context \textit{before} and \textit{after} the action occurs, nor share the clusters between the $2$ pooling layers. As a result, we not only enforce $K = K_a + K_b$, but also set $K_a = K_b = K/2$. Similarly, we set $T_a = T_b = T/2$ and consider the same amount of temporal context from \textit{before} and \textit{after} the actions. We reached highest performances by setting a temporal window $T=15$s and $K=64$ clusters. Finally, we suppress duplicate spottings around the highest confidence score with a NMS considering a centered window of $T_{NMS}=30$s.

\mysection{Training details.} We use the Adam \cite{kingma2014adam} optimizer with default $\beta$ parameters from PyTorch and a starting learning rate of $10^{-3}$ that we decay from a factor of $10$ after the validation loss does not improve for $10$ consecutive epochs. We stop the training once the learning rate decays below $10^{-8}$. Typically, a training converges in ${\sim}100$ epochs corresponding to ${\sim}2$h on a GTX1080Ti with a memory footprint of ${\sim}1$GB. Note that such footprint does not account for the extraction of the ResNet-152 frame features, that were pre-extracted. 
The code is available at \href{https://soccer-net.org/}{https://soccer-net.org/}.

\mysection{Dataset and metrics.} We train our novel architecture with the learnable dimensionality reduction and the temporally-aware pooling module on the SoccerNet dataset using the recent annotations with $17$ classes from \SN2~\cite{SNv2} and the recommended train/val/test split (300/100/100 games).
We consider the action spotting Average-mAP introduced by \SoccerNet, that considers the average precision (AP) for the spotting results per class within a given tolerance $\delta$, averaged per class (mAP). The mAP are further averaged over tolerances ranging from $5$s to $60$s using a step size of $5$s as per common practice \cite{Giancola_2018_CVPR_Workshops,cioppa2020context,SNv2,tomei2021rms}.

\subsection{Main Results}
The main performances of our action spotting architecture based on NetVLAD++ are compared in \Table{MainResults} with the current \sota for action spotting on \SN2. For our method, we report the best model over $5$ runs, as per common practice in video understanding \cite{alwassel2020tsp}, yet report a standard deviation contained within $0.2\%$. The main metric Average-mAP exhibits a boost of $12.7\%$ w.r.t the previous \sota method \CALF. The improvement is consistent across $16$ over the $17$ classes of actions, where only the class \emph{Goal} displayed worst performances. All the methods reported in \Table{MainResults} leverage ResNet-152 features extracted at 2fps (in addition of VGGish audio features for AudioVid \cite{Vanderplaetse2020Improved}). Each of those baselines have different ways to deal with the frame features to solve for action spotting.

\begin{figure}[t]
    \centering
    \includegraphics[width=\linewidth]{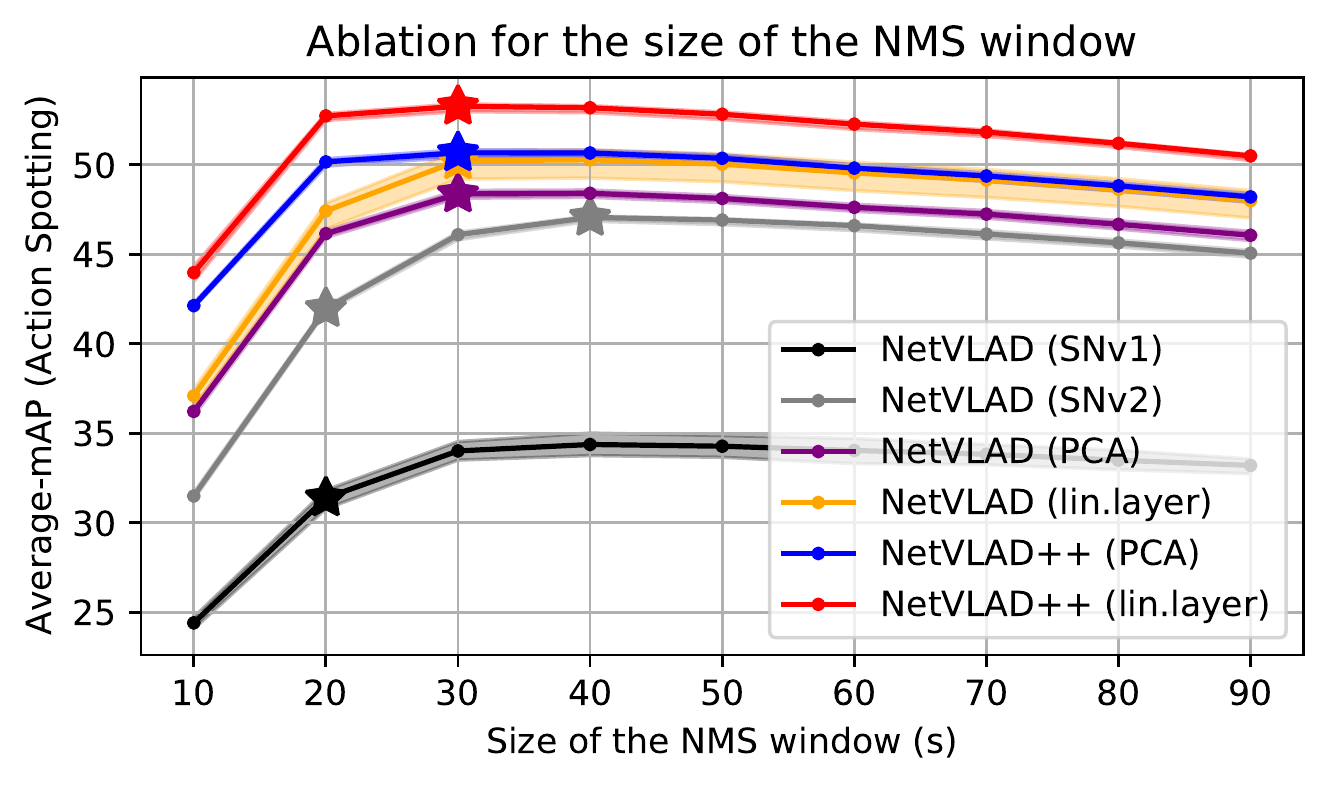}
    \caption{
    \textbf{Ablation on the size of the NMS window.}
    We report the performances of the $4$ models presented in \Table{NetVLADs}, with different size for the NMS window.
    Each entry point is averaged over 5 runs overlaid with max and min performances.
    }
    \label{fig:ablation_wind_NMS}
\end{figure}

\begin{table*}[ht]
\scriptsize
    \caption{\textbf{Ablation Studies.}
    \textbf{(Top) Main Contributions:}
    We highlight the improvement of each component of our novel \textbf{[NetVLAD++]} module and architecture on top of \textbf{[NetVLAD]},  with optimal NMS parameters \textbf{[NMS*]} and optimal window size $T$ \textbf{[NMS*/T*]}.
    We highlight the contribution of the learnable linear layer \textbf{[w/o lin.layer]} and the temporally-aware feature pooling \textbf{[w/o tmp-aware]}.
    All performances are averaged over 5 runs.
    \textbf{(Bottom) Temporal context:} 
    We report the Average-mAP for temporal window size $T$ ranging from $5$ to $30$, averaged over 5 runs. Best performances per class are reported in bold.
    $T=15s$ appears to be optimal.
    }
    \centering
    \setlength{\tabcolsep}{2pt}
    \resizebox{\linewidth}{!}{
    \begin{tabular}{l||c||c|c||c|c|c|c|c|c|c|c|c|c|c|c|c|c|c|c|c}
      &  \begin{turn}{90}\bf SoccerNet-v2\end{turn} &  \begin{turn}{90}visible\end{turn}   &  \begin{turn}{90}unshown\end{turn}  & \begin{turn}{90} Ball out \end{turn} & \begin{turn}{90}Throw-in\end{turn} & \begin{turn}{90}Foul \end{turn} & \begin{turn}{90}Ind. free-kick \end{turn} & \begin{turn}{90}Clearance \end{turn} & \begin{turn}{90}Shots on tar. \end{turn} & \begin{turn}{90}Shots off tar. \end{turn} & \begin{turn}{90}Corner \end{turn} & \begin{turn}{90}Substitution \end{turn} & \begin{turn}{90}Kick-off \end{turn} & \begin{turn}{90}Yellow card \end{turn} & \begin{turn}{90}Offside \end{turn} & \begin{turn}{90}Dir. free-kick \end{turn} & \begin{turn}{90}Goal \end{turn} & \begin{turn}{90}Penalty \end{turn} & \begin{turn}{90}Yel.$\to$Red \end{turn} & \begin{turn}{90}Red card\end{turn} \\ 

       \midrule \midrule
\B Ablation & \multicolumn{20}{c}{\bf Main Contributions } \\ \midrule \midrule
\B NetVLAD          &  31.4 &  34.2 &  23.5 &  46.9 &  41.2 &  31.3 &  17.4 &  34.2 &  18.5 &  19.1 &  55.6 &  50.9 &  46.7 &  31.4 &  17.8 &  34.2 &  54.5 &  33.9 &   0.0 &   0.0  \\ \midrule
\B ~$+$ NMS*      &  47.1 &  52.3 &  34.8 &  60.2 &  57.5 &  52.8 &  38.8 &  54.5 &  36.2 &  36.0 &  72.4 &  66.7 &\B63.4 &  49.6 &  33.0 &  50.6 &  66.3 &  55.0 &   2.4 &   4.5  \\ \midrule
\B ~$+$ NMS*/T*   &  48.4 &  54.2 &  32.5 &  62.8 &  60.0 &  53.8 &  38.8 &  56.2 &  36.6 &  37.7 &  76.7 &  67.2 &  62.5 &  51.8 &  30.8 &  51.2 &  67.0 &  60.2 &   3.8 &   5.2  \\ \midrule\midrule
\B NetVLAD++&\B53.3 &\B59.1 &  35.1 &\B70.2 &\B68.9 &\B64.1 &\B45.2 &\B56.6 &\B38.2 &\B40.4 &\B79.8 &  68.9 &  61.1 &  56.1 &\B38.0 &\B58.2 &\B71.6 &\B79.1 &\B 5.5 &   3.5  \\ \midrule
\B w/o tmp-aware &  50.2 &  56.6 &  32.2 &  64.2 &  61.3 &  54.4 &  39.4 &\B56.6 &  37.3 &  39.6 &  77.6 &  66.1 &  60.7 &\B56.4 &  32.7 &  55.6 &  66.4 &  64.3 &   4.5 &\B16.9  \\ \midrule
\B w/o lin.layer &  50.7 &  55.8 &\B37.3 &  68.2 &  65.3 &  62.4 &  43.4 &  56.0 &  37.1 &  38.3 &  78.9 &\B70.3 &  59.6 &  50.0 &  35.3 &  55.2 &  70.2 &  67.7 &   1.7 &   1.5  \\ \midrule
 \midrule
 
\B Window Size & \multicolumn{20}{c}{\bf Temporal context } \\ \midrule \midrule

\B ~~~$\mathbf{T=05s}$   &  46.0 &  52.8 &  28.6 &  66.9 &  68.6 &  46.5 &  36.1 &  50.2 &  34.9 &\B41.1 &  81.2 &  58.9 &  54.7 &  51.7 &   9.5 &\B58.6 &  45.2 &  64.3 &  12.5 &   1.2  \\ \midrule
\B ~~~$\mathbf{T=10s}$   &  50.7 &  57.1 &  34.5 &\B70.2 &\B70.1 &  61.5 &  42.8 &  53.7 &  37.3 &  39.3 &\B81.9 &  66.6 &  59.1 &  55.1 &  26.8 &  58.3 &  63.1 &  68.6 &   5.8 &   1.7  \\ \midrule
\B ~~~$\mathbf{T=15s}$   &\B53.3 &\B59.1 &\B35.1 &  70.2 &  68.9 &\B64.1 &\B45.2 &\B56.6 &  38.2 &  40.4 &  79.8 &\B68.9 &\B61.1 &\B56.1 &  38.0 &  58.2 &  71.6 &\B79.1 &   5.5 &   3.5  \\ \midrule
\B ~~~$\mathbf{T=20s}$   &  53.0 &  58.3 &  35.1 &  67.5 &  66.1 &  62.2 &  44.5 &  56.4 &  38.5 &  39.5 &  77.2 &  68.9 &  59.1 &  54.8 &  39.0 &  57.0 &\B73.4 &  78.6 &  10.3 &   7.1  \\ \midrule
\B ~~~$\mathbf{T=25s}$   &  50.7 &  55.6 &  34.9 &  64.2 &  62.8 &  59.4 &  45.0 &  55.4 &\B38.9 &  37.3 &  71.0 &  65.2 &  59.6 &  54.5 &  39.1 &  54.0 &  70.9 &  63.7 &\B16.5 &   4.6  \\ \midrule
\B ~~~$\mathbf{T=30s}$   &  49.4 &  53.9 &  35.1 &  60.1 &  57.5 &  54.7 &  43.2 &  51.6 &  38.3 &  35.9 &  65.7 &  62.1 &  59.3 &  55.2 &\B39.4 &  53.8 &  70.8 &  75.5 &   9.4 &\B 7.5  \\ \bottomrule
\end{tabular}
}
    \label{tab:NetVLADs}
\end{table*}

\subsection{Main Ablation Study}
Our improvement originates from $3$ main differences w.r.t NetVLAD: \textbf{(i)} an optimized NMS head to extract spotting results, \textbf{(ii)} a linear layer for the frame features \vs a PCA reduction and \textbf{(iii)} a temporally-aware pooling module. We ablate each component in \Table{NetVLADs} with performances averaged over $5$ runs. \Figure{ablation_wind_NMS} ablates the window size for the NMS and illustrates in transparency the variation between best and worst performances (yet contained).

\mysection{Optimal setup for NetVLAD.} First, we optimized the hyper-parameters for NetVLAD-like pooling methods. In particular, we identified $2$ components that boost further the performances reported in \SN2~\cite{SNv2}: the NMS head and the size of the sliding window $T$. \SoccerNet only considered spotting predictions with confidence scores higher than $0.5$, depicting a lower-bound estimation of the performances ($31.4\%$), as shown in \Figure{ablation_wind_NMS} (\textbf{black}) and \Table{NetVLADs} (\textbf{NetVLAD}). Considering all action spots without the threshold constraint on the confidence score leads to $42.0\%$ Avg-mAP ($+10.6\%$). Yet, further fine-tuning the size of the NMS window leads to $47.1\%$ Avg-mAP ($+5.1\%$), as shown in \Figure{ablation_wind_NMS} (\textbf{\color{Gray}grey}) and \Table{NetVLADs} (\textbf{$\B+$NMS*}). Finally, optimizing the size of the window $T$ from $20s$ to $15s$ leads to an Avg-mAP of $48.4\%$ ($+1.3\%$), as shown in \Figure{ablation_wind_NMS} (\textbf{\color{Purple}purple}) and \Table{NetVLADs} (\textbf{$\B+$NMS*/T*}).

\mysection{NetVLAD++.} 
The learnable linear layer exhibits in \Figure{ablation_wind_NMS} (\textbf{\color{orange}yellow}) and \Table{NetVLADs} (\textbf{w/o tmp-aware}) a $+1.8\%$ boost w.r.t the best optimized NetVLAD ($50.2\%$ Avg-mAP).
Similarly, the temporally-aware pooling exhibit in \Figure{ablation_wind_NMS} (\textbf{\color{Blue}blue}) and \Table{NetVLADs} (\textbf{w/o lin.layer}) a $+2.3\%$ boost w.r.t the best optimized NetVLAD  ($50.7\%$ Avg-mAP).
Our final \textbf{NetVLAD++} displays an Avg-mAP of $53.4\%$, a boost of $4.9\%$ w.r.t to the best optimized NetVLAD (\Figure{ablation_wind_NMS} (\textbf{\color{Red}red})).



\mysection{Ablation per class.} \Table{NetVLADs} further depicts the performances per class. The linear layer mostly improves the performances for the visible instances of actions w.r.t the PCA dimensionality reduction, yet displays a drop for the unshown instances. We argue the unshown actions are diverse hence challenging to learn from. In contrast, the PCA reduction is generic and unsupervised, hence does not not suffer from the challenging input data. Regarding the temporal-awareness, the improvement is consistent regardless of the action visibility.
Most classes appears to provide optimal results with NetVLAD++. Yet, \textit{Clearances}, \textit{Kick-Off}, \textit{Yellow} and \textit{Red cards} appear not to benefit from the temporal-awareness. We hypothesize that that those actions are already discriminative enough without it.
Similarly, \textit{Substitution} and \textit{Kick-Off} do not benefit from the linear projection. Here, we believe the visual cue for those actions are global enough to not require a trainable projection.

\begin{figure}[t]
    \centering
    \includegraphics[width=\linewidth]{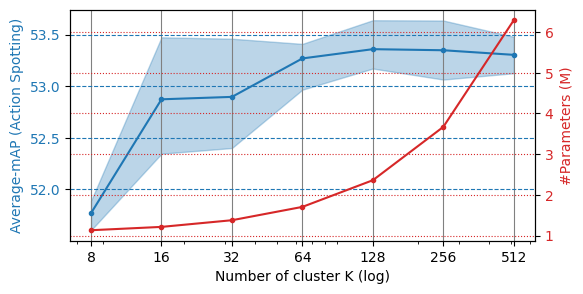}
    \caption{\textbf{Ablation for the Vocabulary Size.}
    The spotting performances are plotted in blue, averaged over 5 runs.
    Light blue illustrate the range of Average-mAP (min/max).
    Red indicated the number of parameters.
    More clusters increase the vocabulary but also the number of parameters, with a saturation after $K=64$.
    }
    \label{fig:AblationK}
\end{figure}

\subsection{Clusters and Temporal Windows}
We further investigate the choice of hyper-parameters for NetVLAD++, in particular the size of the temporal windows $T$ (\Table{NetVLADs}) and the number of cluster $K$ (\Figure{AblationK}). 

For the temporal windows, $T=15$s appears to be optimal for the Average-mAP, yet specific action classes could benefit from different temporal boundaries. We believe that considering a different temporal window per class could lead to optimal results, yet impractical in our architecture that consider a single window to pool features from.

As for the number of clusters, the more the better, yet the performances appears to saturate after $64$ clusters. In fact, $K$ define the size of the vocabulary to cluster the pooled features but the vocabulary can only improve up to a certain extent. Furthermore, note that an increase in the number of cluster irrevocably leads into an increase in the number of parameters. The classifier process the NetVLAD features of dimension $K*D$ in a fully connected layer parameterized with $(K*D+1)$ weight and biases per class. Practically, we refrain on using a large vocabulary $K$ as it leads to large number of parameters and chose $K=64$ in the design of our architecture.

\subsection{More Video Encoders}
Most related works consider ResNet-152 for the video feature encoder~\cite{Giancola_2018_CVPR_Workshops,cioppa2020context,SNv2}. \SoccerNet provides alternative I3D \cite{i3d} and C3D \cite{c3d} video features, yet showed worst performances \cite{Giancola_2018_CVPR_Workshops}. In \Table{MoreEncoder}, we show that our temporally-aware pooling NetVLAD++ transfers well to I3D and C3D, boosting NetVLAD with $6.6\%$ and $2.5\%$, respectfully. More recent video encoders such as R3D \cite{r3d} and R(2+1)D \cite{r25d} could lead to higher performances, but due to the computational complexity of pre-extracting frame features, we leave that for future works.

\begin{table}[t]
    \centering
    \caption{
    \textbf{More video encoder:}
    Spotting performance using I3D, C3D and ResNET-152 video encoders, averaged over 5 runs (means $\pm$ std).
    NetVLAD++ with linear layer for dimensionality reduction results in best performances for all encoders.
    }
    \begin{tabular}{c||c|c|c}
Pooling & NetVLAD & NetVLAD++ & NetVLAD++ \\ \midrule 
Encoder & (PCA) & (PCA) & (lin layer) \\ \midrule \midrule 
I3D & $34.9  \pm 0.3$  & $38.1  \pm 0.1$  & $ \mathbf { 41.5  \pm 0.1 } $ \\ \midrule 
C3D & $46.1  \pm 0.3$  & $47.2  \pm 0.2$  & $ \mathbf { 48.6  \pm 0.8 } $ \\ \midrule 
ResNet & $48.4  \pm 0.2$  & $50.7  \pm 0.2$  & $ \mathbf { 53.3  \pm 0.2 } $ \\ \bottomrule 
    \end{tabular}
    \label{tab:MoreEncoder}
\end{table}

\subsection{More Temporally-Aware Pooling Modules}
We further transfer our temporally-aware pooling to further pooling modules. In particular, we implement temporally-aware Max, Average and NetRVLAD pooling modules as reported in \Table{MorePooling}. We refrained in optimizing the performances for each module, and only highlight the relative improvement brought by the temporal awareness.

\begin{table}[t]
    \centering
    \caption{\textbf{More pooling modules.}
    Spotting performances using Max, Avg and NetRVLAD pooling modules.
    All temporally-aware pooling method outperforms the original pooling. 
    }
    \begin{tabular}{c||c|c}

Pooling & Original & Tmp.-Aware \\ \midrule  \midrule
MaxPooling (PCA)     & $23.7  \pm 0.4$  & $ \mathbf { 31.6  \pm 0.7 } $ \\ \midrule 
AvgPooling (PCA)     & $32.5  \pm 0.1$  & $ \mathbf { 40.6  \pm 0.2 } $ \\ \midrule 
NetRVLAD (lin.layer) & $48.0  \pm 0.2$  & $ \mathbf { 50.9  \pm 0.3 } $ \\ \midrule 
NetVLAD  (lin.layer) & $50.2  \pm 0.6$  & $ \mathbf { 53.3  \pm 0.2 } $ \\ \bottomrule 
    \end{tabular}
    \label{tab:MorePooling}
\end{table}

We developed MaxPool++ and AvgPool++ based on MaxPool and AvgPool with an extra temporal awareness. We considered the PCA-reduced ResNet features as the low number of parameters for those models ($9234$ parameters each) impeded a stable learning on top of higher dimensionality features. Note that MaxPool++ and AvgPool++ concatenates the \textit{past} and \textit{future} context (twice the dimensionality) which inevitably leads to a similar increase in parameters ($18450$ parameters each). 
MaxPool++ and AvgPool++ displayed impressive boosts in performances ($+7.9\%$ and $+8.1\%$ resp.) now flirting with performances similar to previous baselines proposed in \SN2~\cite{SNv2}, yet leveraging ${\sim}20\times$ less parameters for the spotting head.

Following previous experiments on residual-less NetVLAD \cite{Giancola_2018_CVPR_Workshops}, we developed NetRVLAD++ on top of NetRVLAD, which drops the cluster parameters $\cl_k(j)$ in \eqref{eq:netvlad}, leading to slightly less parameters to learn. We build NetRVLAD++ on top of full ResNet feature with our learnable feature projection. The relative improvement here is similar ($+3.1\%$), yet the performances are not on par ($-2.2\%$) with NetVLAD++, highlighting the importance of the NetVLAD residuals.

\section{Discussion}

\mysection{Temporal-awareness \vs \CALF \vs \RMS.}
NetVLAD++ is not the first approach that considers temporal semantic regions around the action to spot. 
\CALF defines a high-level semantic context from different temporal regions \textit{far distant}, \textit{just before} and \textit{just after} an action occurs. 
They introduce a hand-crafted loss function that weights the contextual information.
Still, they leverage the same features for the context \textit{before} and \textit{after} the actions occurs. 
In contrast, we drop the \textit{far distant} semantic context in NetVLAD++ and learn specific features from different vocabulary (NetVLAD clusters) for the \textit{past} and \textit{future} temporal context.
\RMS propose a similar contextual approach borrowed from the NLP literature that masks out part of the temporal context.
In particular, they drop the \textit{past} information during training, expecting the model to focus exclusively on the future frames.
In contrast, we learn both \textit{past} and \textit{future} temporal context independently on NetVLAD++, and merge both learned context.

\mysection{More temporal regions.} We considered extending the temporal region beyond the close \textit{past} and \textit{future} contexts, following insights from \CALF that considered \textit{far distant} temporal segments.
Our experiments with \textit{far before} and \textit{far after} temporal contexts did not lead to any improvement for the learning of the pooling module and inevitably increases the number of hyper-parameters defining those temporal regions. We believe the temporal context \textit{before} and \textit{after} are discriminative enough, while the \textit{far distant} equivalent are more blurry in time with the \textit{close} context. 
Also, each action class might consider different temporal context for the \textit{far distant}, which would lead to more confusion for the learnable pooling layers. 
We believe a global video feature or a better temporal aggregation of the features across the complete video could lead to a better temporal understanding and would take care of the \textit{far distant} temporal context.

\mysection{Spotting Regression.}
Both \CALF and \RMS learn to regress action spots. We decided not to regress the actions spot but rather rely on a dense sliding windows with an NMS to discard non-optimal action spots. A dense sliding window inevitably leads to slower inference, yet NetVLAD++ takes ${<}1$ second to infer a complete $90$min soccer game from pre-extracted features.

\mysection{\SN2 Challenge.}
We tested our approach on the segregated challenge set of \SN2. For the competition, we trained on the \textit{train+val} sets, validated on the \textit{test} set and inferred on the \textit{challenge} set, that we submitted on the evaluation server. At submission time (\Table{challenge}), we reached SOTA performances with $52.54\%$ Avg-mAP.


\begin{table}[t]
    \centering
    \caption{
    \textbf{\SN2 Challenge.}
    Our NetVLAD++ approach reach best performances on the SoccerNet
    }
    \begin{tabular}{l||c||c|c}
        Method & Avg-mAP & Visible & Unshown \\ \midrule \midrule
        NetVLAD \cite{Giancola_2018_CVPR_Workshops} & 30.74 & 32.99 & 23.27 \\ \midrule
        \CALF & 42.22 & 43.51 & 37.91 \\ \midrule
        \RMS & 49.66 & 53.11 & 38.92 \\ \midrule \midrule
        NetVLAD++ & \B 52.54 & \B57.12 &\B46.15 \\ \bottomrule
    \end{tabular}
    \label{tab:challenge}
\end{table}


\section{Conclusion}
\label{sec:Conclusion}


In this work, we proposed a temporally-aware learnable pooling module for the task of action spotting on soccer videos. 
We first showed that NetVLAD can further be optimized on \SN2.
%
We further improve the pooling-based action spotting architecture by learning a linear projection that reduce the features dimension and split the past and future features to pool, leading to \sota performances on the \SN2 benchmark.
We show a complete ablation and transfer capability of our contribution to any pooling layer and input features, paving the road for more temporally-aware learning in video.
We believe future works should focus on integrating local frame features from low-level semantics (player, ball, field, etc...) and consider complete videos rather than temporally-bounded clips as input.
Future works should learn to accumulate knowledge in time or based on attention models in order to reach higher-level of understanding in soccer broadcasts.

\small{
\mysection{Acknowledgments:}
This work is supported by the KAUST Office of Sponsored Research under Award No. OSR-CRG2017-3405.}









\clearpage

{\small
\bibliographystyle{ieee_fullname}
\bibliography{bibliography,biblio_CALF,biblio_SoccerNet}

\begin{thebibliography}{10}\itemsep=-1pt

\bibitem{alwassel2020tsp}
Humam Alwassel, Silvio Giancola, and Bernard Ghanem.
\newblock Tsp: Temporally-sensitive pretraining of video encoders for
  localization tasks.
\newblock {\em arXiv preprint arXiv:2011.11479}, 2020.

\bibitem{arandjelovic2016netvlad}
Relja Arandjelovic, Petr Gronat, Akihiko Torii, Tomas Pajdla, and Josef Sivic.
\newblock {NetVLAD}: {CNN} architecture for weakly supervised place
  recognition.
\newblock In {\em CVPR}, pages 5297--5307, 2016.

\bibitem{arandjelovic2013all}
Relja Arandjelovic and Andrew Zisserman.
\newblock All about {VLAD}.
\newblock In {\em CVPR}, pages 1578--1585, 2013.

\bibitem{bridgeman2019multi}
Lewis Bridgeman, Marco Volino, Jean-Yves Guillemaut, and Adrian Hilton.
\newblock Multi-person 3d pose estimation and tracking in sports.
\newblock In {\em Proceedings of the IEEE/CVF Conference on Computer Vision and
  Pattern Recognition Workshops}, pages 0--0, 2019.

\bibitem{Bridgeman_2019_CVPR_Workshops}
Lewis Bridgeman, Marco Volino, Jean-Yves Guillemaut, and Adrian Hilton.
\newblock {Multi-Person 3D Pose Estimation and Tracking in Sports}.
\newblock In {\em IEEE Conference on Computer Vision and Pattern Recognition
  (CVPR) Workshops}, pages 2487--2496, June 2019.

\bibitem{i3d}
Joao Carreira and Andrew Zisserman.
\newblock Quo vadis, action recognition? a new model and the kinetics dataset.
\newblock In {\em proceedings of the IEEE Conference on Computer Vision and
  Pattern Recognition}, pages 6299--6308, 2017.

\bibitem{cioppa2020context}
Anthony Cioppa, Adrien Deli{\`e}ge, Silvio Giancola, Bernard Ghanem, Marc~Van
  Droogenbroeck, Rikke Gade, and Thomas~B. Moeslund.
\newblock A context-aware loss function for action spotting in soccer videos.
\newblock In {\em IEEE Conference on Computer Vision and Pattern Recognition
  (CVPR)}, pages 13126--13136, 2020.

\bibitem{Cioppa_2019_CVPR_Workshops}
Anthony Cioppa, Adrien Deli{\`e}ge, Maxime Istasse, Christophe De~Vleeschouwer,
  and Marc Van~Droogenbroeck.
\newblock {ARTHuS: Adaptive Real-Time Human Segmentation in Sports Through
  Online Distillation}.
\newblock In {\em IEEE Conference on Computer Vision and Pattern Recognition
  (CVPR) Workshops}, pages 2505--2514, June 2019.

\bibitem{Cioppa2018ABottom}
Anthony Cioppa, Adrien Deli{\`e}ge, and Marc Van~Droogenbroeck.
\newblock A bottom-up approach based on semantics for the interpretation of the
  main camera stream in soccer games.
\newblock In {\em IEEE Conference on Computer Vision and Pattern Recognition
  (CVPR) Workshops}, pages 1846--1855, June 2018.

\bibitem{SNv2}
Adrien Deli{\`e}ge, Anthony Cioppa, Silvio Giancola, Meisam~J Seikavandi,
  Jacob~V Dueholm, Kamal Nasrollahi, Bernard Ghanem, Thomas~B Moeslund, and
  Marc Van~Droogenbroeck.
\newblock Soccernet-v2: A dataset and benchmarks for holistic understanding of
  broadcast soccer videos.
\newblock {\em arXiv preprint arXiv:2011.13367}, 2020.

\bibitem{SportsBroadcastUS}
Deloitte.
\newblock Length of sports tv broadcast hours in the united states from 2002 to
  2017.
\newblock In {\em Statista - The Statistics Portal}, 2021.
\newblock Retrieved January 11, 2021, from
  \url{https://www.statista.com/statistics/290110/length-sports-tv-programming-available-usa/}.

\bibitem{SportsMediaRightsUS}
Deloitte.
\newblock Sports media rights market size in north america from 2006 to 2023
  (in billion u.s. dollars)*.
\newblock In {\em Statista - The Statistics Portal}, 2021.
\newblock Retrieved January 11, 2021, from
  \url{https://www.statista.com/statistics/194225/sports-media-rights-revenue-in-north-america/}.

\bibitem{VolumeSportsFrance}
Deloitte.
\newblock Volume of sports programs on free television in france between 2010
  and 2018.
\newblock In {\em Statista - The Statistics Portal}, 2021.
\newblock Retrieved January 11, 2021, from
  \url{https://www.statista.com/statistics/1016830/sports-television-hourly-volume-france/}.

\bibitem{VolumeSportsFrancePay}
Deloitte.
\newblock Volume of sports programs on pay television in france between 2000
  and 2018.
\newblock In {\em Statista - The Statistics Portal}, 2021.
\newblock Retrieved January 11, 2021, from
  \url{https://www.statista.com/statistics/1025759/hourly-volume-sports-pay-tv-france/}.

\bibitem{deng2009imagenet}
Jia Deng, Wei Dong, Richard Socher, Li-Jia Li, Kai Li, and Li Fei-Fei.
\newblock Imagenet: A large-scale hierarchical image database.
\newblock In {\em CVPR}, pages 248--255. IEEE, 2009.

\bibitem{devlin2018bert}
Jacob Devlin, Ming-Wei Chang, Kenton Lee, and Kristina Toutanova.
\newblock Bert: Pre-training of deep bidirectional transformers for language
  understanding.
\newblock {\em arXiv preprint arXiv:1810.04805}, 2018.

\bibitem{farin2003robust}
Dirk Farin, Susanne Krabbe, Peter de With, and Wolfgang Effelsberg.
\newblock {Robust camera calibration for sport videos using court models}.
\newblock In {\em Storage and Retrieval Methods and Applications for
  Multimedia}, pages 80--91, December 2003.

\bibitem{felsen2017will}
Panna Felsen, Pulkit Agrawal, and Jitendra Malik.
\newblock {What will happen next? Forecasting player moves in sports videos}.
\newblock In {\em IEEE International Conference on Computer Vision (ICCV)},
  pages 3362--3371, October 2017.

\bibitem{Giancola_2018_CVPR_Workshops}
Silvio Giancola, Mohieddine Amine, Tarek Dghaily, and Bernard Ghanem.
\newblock {SoccerNet: A Scalable Dataset for Action Spotting in Soccer Videos}.
\newblock In {\em IEEE Conference on Computer Vision and Pattern Recognition
  (CVPR) Workshops}, pages 1711--1721, June 2018.

\bibitem{he2016deep}
Kaiming He, Xiangyu Zhang, Shaoqing Ren, and Jian Sun.
\newblock Deep residual learning for image recognition.
\newblock In {\em IEEE International Conference on Computer Vision and Pattern
  Recognition (CVPR)}, pages 770--778, June 2016.

\bibitem{heidarivincheh2017detecting}
Farnoosh Heidarivincheh, Majid Mirmehdi, and Dima Damen.
\newblock Detecting the moment of completion: temporal models for localising
  action completion.
\newblock {\em arXiv preprint arXiv:1710.02310}, 2017.

\bibitem{caba2015activitynet}
Fabian~Caba Heilbron, Victor Escorcia, Bernard Ghanem, and Juan~Carlos Niebles.
\newblock {ActivityNet: A Large-Scale Video Benchmark for Human Activity
  Understanding}.
\newblock In {\em IEEE International Conference on Computer Vision and Pattern
  Recognition (CVPR)}, pages 961--970, June 2015.

\bibitem{homayounfar2017sports}
Namdar Homayounfar, Sanja Fidler, and Raquel Urtasun.
\newblock {Sports field localization via deep structured models}.
\newblock In {\em IEEE International Conference on Computer Vision and Pattern
  Recognition (CVPR)}, pages 4012--4020, July 2017.

\bibitem{Istasse_2019_CVPR_Workshops}
Maxime Istasse, Julien Moreau, and Christophe De~Vleeschouwer.
\newblock {Associative Embedding for Team Discrimination}.
\newblock In {\em IEEE Conference on Computer Vision and Pattern Recognition
  (CVPR) Workshops}, pages 2477--2486, June 2019.

\bibitem{jegou2010aggregating}
Herv{\'e} J{\'e}gou, Matthijs Douze, Cordelia Schmid, and Patrick P{\'e}rez.
\newblock Aggregating local descriptors into a compact image representation.
\newblock In {\em CVPR}, pages 3304--3311. IEEE, 2010.

\bibitem{Jiang2020SoccerDB}
Yudong Jiang, Kaixu Cui, Leilei Chen, Canjin Wang, and Changliang Xu.
\newblock Soccerdb: A large-scale database for comprehensive video
  understanding.
\newblock In {\em International Workshop on Multimedia Content Analysis in
  Sports}, pages 1--8, October 2020.

\bibitem{kingma2014adam}
Diederik~P Kingma and Jimmy Ba.
\newblock Adam: A method for stochastic optimization.
\newblock {\em arXiv preprint arXiv:1412.6980}, 2014.

\bibitem{manafifard2017survey}
Mehrtash Manafifard, Hamid Ebadi, and Hamid~Abrishami Moghaddam.
\newblock {A survey on player tracking in soccer videos}.
\newblock {\em Computer Vision and Image Understanding}, 159:19--46, June 2017.

\bibitem{moeslund2014computer}
Thomas~B. Moeslund, Graham Thomas, and Adrian Hilton.
\newblock {\em {Computer Vision in Sports}}.
\newblock Springer, 2014.

\bibitem{Rematas_2018_CVPR}
Konstantinos Rematas, Ira Kemelmacher-Shlizerman, Brian Curless, and Steve
  Seitz.
\newblock Soccer on your tabletop.
\newblock In {\em IEEE International Conference on Computer Vision and Pattern
  Recognition (CVPR)}, pages 4738--4747, June 2018.

\bibitem{ren2016faster}
Shaoqing Ren, Kaiming He, Ross Girshick, and Jian Sun.
\newblock Faster r-cnn: towards real-time object detection with region proposal
  networks.
\newblock {\em IEEE transactions on pattern analysis and machine intelligence},
  39(6):1137--1149, 2016.

\bibitem{rongved-ism2020}
Olav A.~Nerg{\aa}rd Rongved, Steven~A. Hicks, Vajira Thambawita, H{\aa}kon~K.
  Stensland, Evi Zouganeli, Dag Johansen, Michael~A. Riegler, and P{\aa}l
  Halvorsen.
\newblock Real-time detection of events in soccer videos using {3D}
  convolutional neural networks.
\newblock In {\em IEEE International Symposium on Multimedia (ISM)}, December
  2020. In press.

\bibitem{Sangesa2020UsingPB}
Adri{\`a}~Arbu{\'e}s Sang{\"u}esa, A. Mart{\'i}n, J. Fern{\'a}ndez, C.
  Ballester, and G. Haro.
\newblock Using player's body-orientation to model pass feasibility in soccer.
\newblock In {\em IEEE Conference on Computer Vision and Pattern Recognition
  (CVPR) Workshops}, pages 3875--3884, June 2020.

\bibitem{Sarkar_2019_CVPR_Workshops}
Saikat Sarkar, Amlan Chakrabarti, and Dipti Prasad~Mukherjee.
\newblock {Generation of Ball Possession Statistics in Soccer Using
  Minimum-Cost Flow Network}.
\newblock In {\em IEEE Conference on Computer Vision and Pattern Recognition
  (CVPR) Workshops}, pages 2515--2523, June 2019.

\bibitem{Theagarajan_2018_CVPR_Workshops}
Rajkumar Theagarajan, Federico Pala, Xiu Zhang, and Bir Bhanu.
\newblock Soccer: Who has the ball? {Generating} visual analytics and player
  statistics.
\newblock In {\em IEEE Conference on Computer Vision and Pattern Recognition
  (CVPR) Workshops}, pages 1830--1838, June 2018.

\bibitem{thomas2017computer}
Graham Thomas, Rikke Gade, Thomas~B. Moeslund, Peter Carr, and Adrian Hilton.
\newblock {Computer vision for sports: Current applications and research
  topics}.
\newblock {\em Computer Vision and Image Understanding}, 159:3--18, June 2017.

\bibitem{tomei2021rms}
Matteo Tomei, Lorenzo Baraldi, Simone Calderara, Simone Bronzin, and Rita
  Cucchiara.
\newblock Rms-net: Regression and masking for soccer event spotting.
\newblock {\em arXiv preprint arXiv:2102.07624}, 2021.

\bibitem{c3d}
Du Tran, Lubomir Bourdev, Rob Fergus, Lorenzo Torresani, and Manohar Paluri.
\newblock Learning spatiotemporal features with 3d convolutional networks.
\newblock In {\em Proceedings of the IEEE international conference on computer
  vision}, pages 4489--4497, 2015.

\bibitem{tran2018closer}
Du Tran, Heng Wang, Lorenzo Torresani, Jamie Ray, Yann LeCun, and Manohar
  Paluri.
\newblock A closer look at spatiotemporal convolutions for action recognition.
\newblock In {\em Proceedings of the IEEE conference on Computer Vision and
  Pattern Recognition}, pages 6450--6459, 2018.

\bibitem{r25d}
Du Tran, Heng Wang, Lorenzo Torresani, Jamie Ray, Yann LeCun, and Manohar
  Paluri.
\newblock A closer look at spatiotemporal convolutions for action recognition.
\newblock In {\em Proceedings of the IEEE conference on Computer Vision and
  Pattern Recognition}, pages 6450--6459, 2018.

\bibitem{Vanderplaetse2020Improved}
Bastien Vanderplaetse and Stephane Dupont.
\newblock Improved soccer action spotting using both audio and video streams.
\newblock In {\em IEEE Conference on Computer Vision and Pattern Recognition
  (CVPR) Workshops}, pages 3921--3931, June 2020.

\bibitem{vats2020event}
Kanav Vats, Mehrnaz Fani, Pascale Walters, David~A Clausi, and John Zelek.
\newblock Event detection in coarsely annotated sports videos via parallel
  multi-receptive field 1d convolutions.
\newblock In {\em IEEE Conference on Computer Vision and Pattern Recognition
  (CVPR) Workshops}, pages 882--883, June 2020.

\bibitem{r3d}
Huijuan Xu, Abir Das, and Kate Saenko.
\newblock R-c3d: Region convolutional 3d network for temporal activity
  detection.
\newblock In {\em Proceedings of the IEEE international conference on computer
  vision}, pages 5783--5792, 2017.

\bibitem{yang2017robust}
Ying Yang and Danyang Li.
\newblock {Robust player detection and tracking in broadcast soccer video based
  on enhanced particle filter}.
\newblock {\em Journal of Visual Communication and Image Representation},
  46:81--94, July 2017.

\bibitem{yu2018comprehensive}
Junqing Yu, Aiping Lei, Zikai Song, Tingting Wang, Hengyou Cai, and Na Feng.
\newblock Comprehensive dataset of broadcast soccer videos.
\newblock In {\em 2018 IEEE Conference on Multimedia Information Processing and
  Retrieval (MIPR)}, pages 418--423. IEEE, 2018.

\bibitem{Zecha_2019_CVPR_Workshops}
Dan Zecha, Moritz Einfalt, and Rainer Lienhart.
\newblock Refining joint locations for human pose tracking in sports videos.
\newblock In {\em IEEE Conference on Computer Vision and Pattern Recognition
  (CVPR) Workshops}, pages 2524--2532, June 2019.

\end{thebibliography}
}

\clearpage 

\newpage 

\end{document}